\pdfoutput=1
\documentclass[11pt,a4paper]{article}
\usepackage[hyperref]{acl2021}
\usepackage{times}
\usepackage{latexsym}
\usepackage{amsmath}
\usepackage{caption}
\usepackage{subcaption}
\usepackage{multirow}
\usepackage{color}
\usepackage{threeparttable}
\usepackage{graphicx}
\usepackage{xspace}
\usepackage{bm}
\usepackage{xfrac}
\usepackage{array}
\usepackage{threeparttable}
\usepackage{algorithm}
\usepackage{algorithmic}
\usepackage{makecell}
\usepackage{balance}
\usepackage{microtype}
\usepackage[symbol]{footmisc}

\usepackage{microtype}

\aclfinalcopy 
\def\aclpaperid{806} 

\newcommand{\oea}{\textsf{OntoEA}\xspace}

\title{OntoEA: Ontology-guided Entity Alignment \\ via Joint Knowledge Graph Embedding}

\author{
  {\spaceskip=0.15em\relax
  Yuejia Xiang$^{1,2}$\footnotemark[1] , Ziheng Zhang$^1$\footnotemark[1] , Jiaoyan Chen$^3$, Xi Chen$^{1,2}$\footnotemark[2] , Zhenxi Lin$^1$, Yefeng Zheng$^1$}
  \\
  $^1$Tencent Jarvis Lab, Shenzhen, China \\
  $^2$Platform and Content Group, Tencent, Shenzhen, China \\
  $^3$Department of Computer Science, University of Oxford, UK \\
  {\normalsize \tt \{yuejiaxiang,zihengzhang\}@tencent.com, jiaoyan.chen@cs.ox.ac.uk} \\
  {\normalsize \tt \{jasonxchen,chalerislin,yefengzheng\}@tencent.com}\\
}

\begin{document}

\maketitle

\footnotetext[1]{Equal contribution}
\footnotetext[2]{Corresponding author}
\renewcommand{\thefootnote}{\arabic{footnote}}

\begin{abstract}
Semantic embedding has been widely investigated for aligning knowledge graph (KG) entities.
Current methods have explored and utilized the graph structure, the entity names, and attributes, but ignore the ontology (or ontological schema) which contains critical meta information such as classes and their membership relationships with entities.
In this paper, we propose an ontology-guided entity alignment method named \oea, where both KGs and their ontologies are jointly embedded, and the class hierarchy and the class disjointness are utilized to avoid false mappings.
Extensive experiments on seven public and industrial benchmarks have demonstrated the state-of-the-art performance of \oea and the effectiveness of the ontologies.

\end{abstract}

\section{Introduction}
Knowledge graphs (KGs) that are composed of entities and facts in the RDF form (i.e., RDF triples) are of vital importance in various applications, such as search engines and personal assistants~\cite{hogan2020knowledge}.
Although there have been several large-scale KGs, the content of one individual KG is often incomplete, especially in supporting some domain-specific applications such as clinical AI assistants.
As these KGs are developed separately, they are usually heterogeneous and supplementary to each other.
Thus it becomes urgently needed to align multiple KGs (i.e., matching the equivalent elements) to fully explore their usability.

\begin{figure}[ht]
\centering
\includegraphics[width=0.95\linewidth]{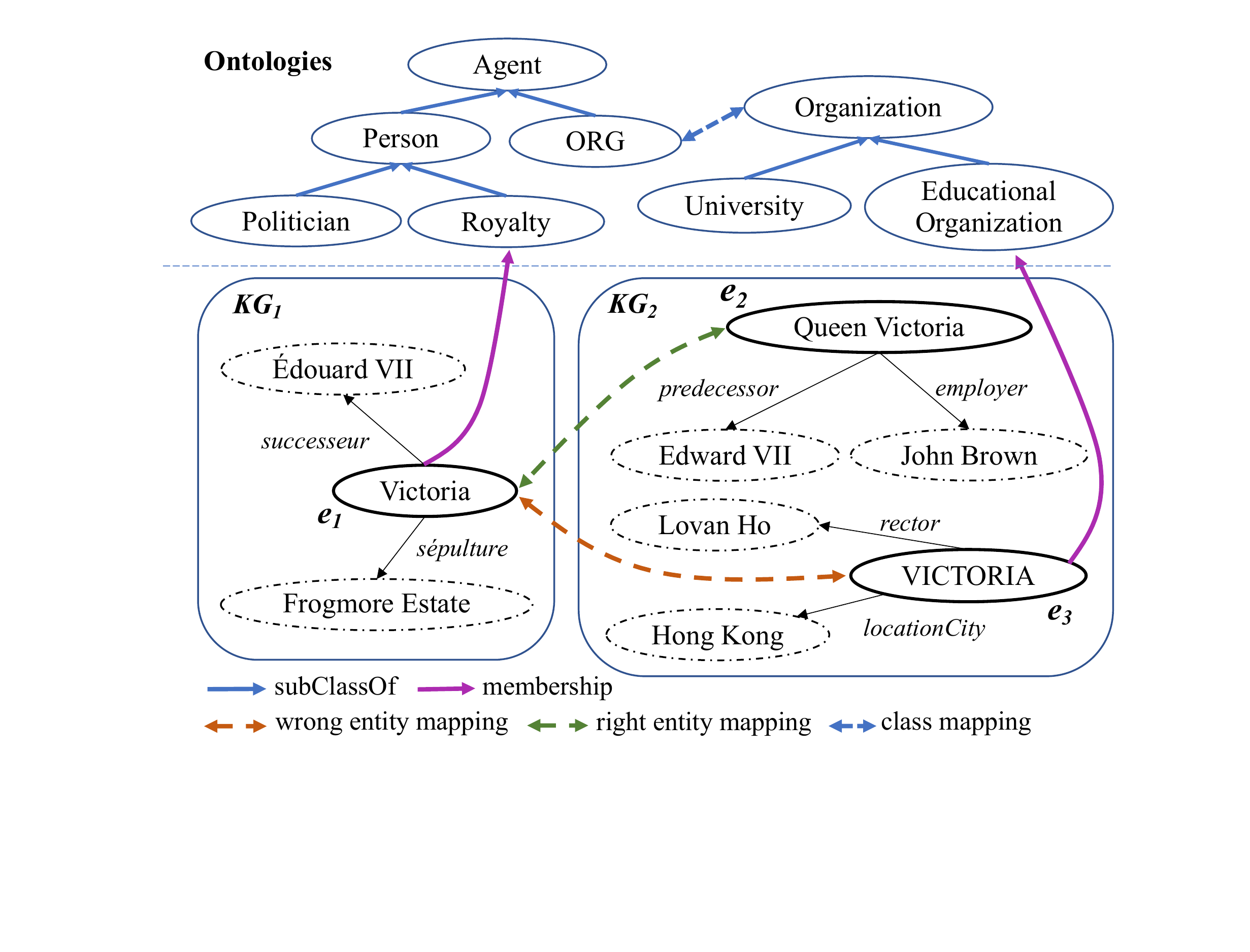}
\caption{A class conflict example in entity alignment.}
\label{figure:intro}
\vspace{-5mm}
\end{figure}

Recently, a few embedding-based methods have been proposed for entity alignment (EA)~\cite{sun2020benchmarking, zhang2020industry, qi2021unsupervised}, in which entities are embedded into vectors and the equivalent entities are determined via calculating the similarity of their vectors.
They extend the embedding of one KG to the embedding of multiple KGs into one vector space by, for instance, a transformation matrix, which is learnt from annotated mappings (a.k.a. seed mappings).
They also utilize the entity names and attributes besides the graph structures for better performance.
These methods, however, all ignore the ontology (or ontological schema) which is an important part of many KGs such as DBpedia~\cite{auer2007dbpedia} and Wikidata \cite{vrandevcic2014wikidata} as meta information for higher quality and usability.
The ontology typically contains hierarchical classes and properties, and optionally defines some logical constraints such as the class disjointness, the property domain and range~\cite{horrocks2008ontologies}.
Meanwhile, the KG usually clarifies the membership relationship between entities and classes.

The entity mappings by the above methods that do not exploit the ontologies may induce some class conflicts.
Considering the example in Fig.~\ref{figure:intro}, \textit{Victoria} in $\text{KG}_{\text{1}}$ is often incorrectly aligned to \textit{VICTORIA} in $\text{KG}_{\text{2}}$ by many embedding-based methods, but they belong to two potentially disjointed classes, namely \textit{Person} and \textit{Organization}.
We find such class conflicted mappings are quite common in the wrongly predicted mappings.
Considering the EN-FR-15K-V1 benchmark by \cite{sun2020benchmarking}, $42.2\%$ and $55.7\%$ of the wrongly predicted mappings are class conflicted when running BootEA~\cite{sun2018bootstrapping} and RSN4EA~\cite{guo2019learning}, respectively.
These false positive mappings could be avoided if their ontologies are considered.

In this study, we propose an ontology-guided entity alignment method named \oea, which enriches the embedded semantics and avoids wrong mappings by exploring the class conflicts. 
\oea can work well in two contexts:  \textit{(i)} one ontology is shared across the to-be-aligned KGs; \textit{(ii)} the ontologies of the to-be-aligned KGs are separated. 
The first context seems to follow a strict assumption but is actually quite common.
For example, DBpedia has multilingual KGs (or versions) that share the same ontology, and in industrial scenarios, it is preferred to create a common ontology and then construct different KGs from different sources.
The second context can be transformed into the first context by pre-aligning the ontologies, using existing ontology alignment systems such as PARIS~\cite{suchanek2011paris} and LogMap~\cite{jimenez2011logmap}, and/or cost-sensitive human intervention.

There are two challenges to utilize the ontology.
First, it is difficult to embed two KGs together with the ontology, and as far as we know, there are currently no such solutions.
Second, the class conflicts, indicated by the class disjointness in the ontology, are not all explicitly defined.
Most of them should be learned from the KGs and they actually vary from KG to KG.
For example, two classes \textit{Human} and \textit{Animal} are often disjointed in artwork KGs, but \textit{Human} may belong to \textit{Animal} in biological KGs.
To address these challenges, we develop a joint embedding method that includes five modules for embedding the KGs, the ontology, the class conflicts, the membership relationships, and the seed mappings, respectively.
Specifically, we develop a class conflict matrix (CCM) to represent different kinds of class conflicts, including those explicitly defined as disjoint, those indicated by the class hierarchy (e.g., sibling classes likely to be disjointed) and those plausible deduced from the KGs.

To the best of our knowledge, \oea is the first to utilize the ontology and the embedding for KG alignment.
Extensive experiments on seven benchmarks have verified the effectiveness of the ontology guidance with both one shared ontology or two separated ontologies.
\oea consistently outperforms the state-of-the-art baselines on all the benchmarks.
For example, it on average achieves over $35\%$ higher Hits@1, Hits@5, and MRR than the best baseline on MED-BBK-9K --- a new and challenging industrial benchmark.
Last but not least, we extend the current benchmarks with ontologies and membership relationships.
The source code and benchmarks are publicly accessible at \url{https://github.com/ZihengZZH/OntoEA}.

\section{Methodology}
\subsection{Problem Statement}

A KG is denoted as $G=(E,R,T)$, where $E,R,T$ are the sets of entities, relations and triples, respectively. 
Each triple $(h,r,t) \in T$, includes a head entity $h \in E$, a relation $r \in R$, and a tail entity $t \in E$.
Their embeddings are denoted as $\bm{h}$, $\bm{r}$, $\bm{t}$, respectively.
For two to-be-aligned KGs $G_i=(E_i,R_i,T_i)$ and $G_j=(E_j,R_j,T_j)$, an entity mapping, denoted as $m=(e_i, e_j)$ where $e_i \in E_i$, $e_j \in E_j$, indicates that $e_i$ and $e_j$ refer to the same real-world object.
The entity alignment (EA) task aims to find all the mappings $M$ between $E_i$ and $E_j$, where we assume a small set of known entity mappings (or seed mappings) $M_{s}$ are given.

Each KG is assumed to be associated with an ontology, which contains hierarchical classes and optionally disjoint constraints between classes.
For simplicity, we consider the classes as entities in KG and the subsumption relationships between classes, often known as \textit{rdfs:subClassOf}, as relations in KG. The simplified ontology is therefore regarded as a graph.
For KGs $G_i$ and $G_j$, their associated ontologies are represented as $O_i = (C_i, H_i)$ and $O_j = (C_j, H_j)$, respectively.
We simply denote both $O_i$ and $O_j$ as $O=(C, H)$ if $G_i$ and $G_j$ share one ontology (i.e., $O_i \equiv O_j$); otherwise we denote the merged ontology after ontology alignment as $O=(C, H)$.
Note $C_i$, $C_j$ and $C$ are the class sets, while $H_i$, $H_j$ and $H$ are the triple sets with only the \textit{subClassOf} relation.
Furthermore, the membership relationships, which link the entities and the corresponding classes, are denoted as $B_i$ and $B_j$, and e.g., $B_i$ links $G_i$ and $C$ via $b_i=(e_i,c)$ where $e_i \in E_i$ and $c \in C$.
The class and membership embeddings are denoted as $\bm{c}$ and $\bm{b}$, respectively.

\begin{figure}[ht]
\centering
\includegraphics[width=1.0\linewidth]{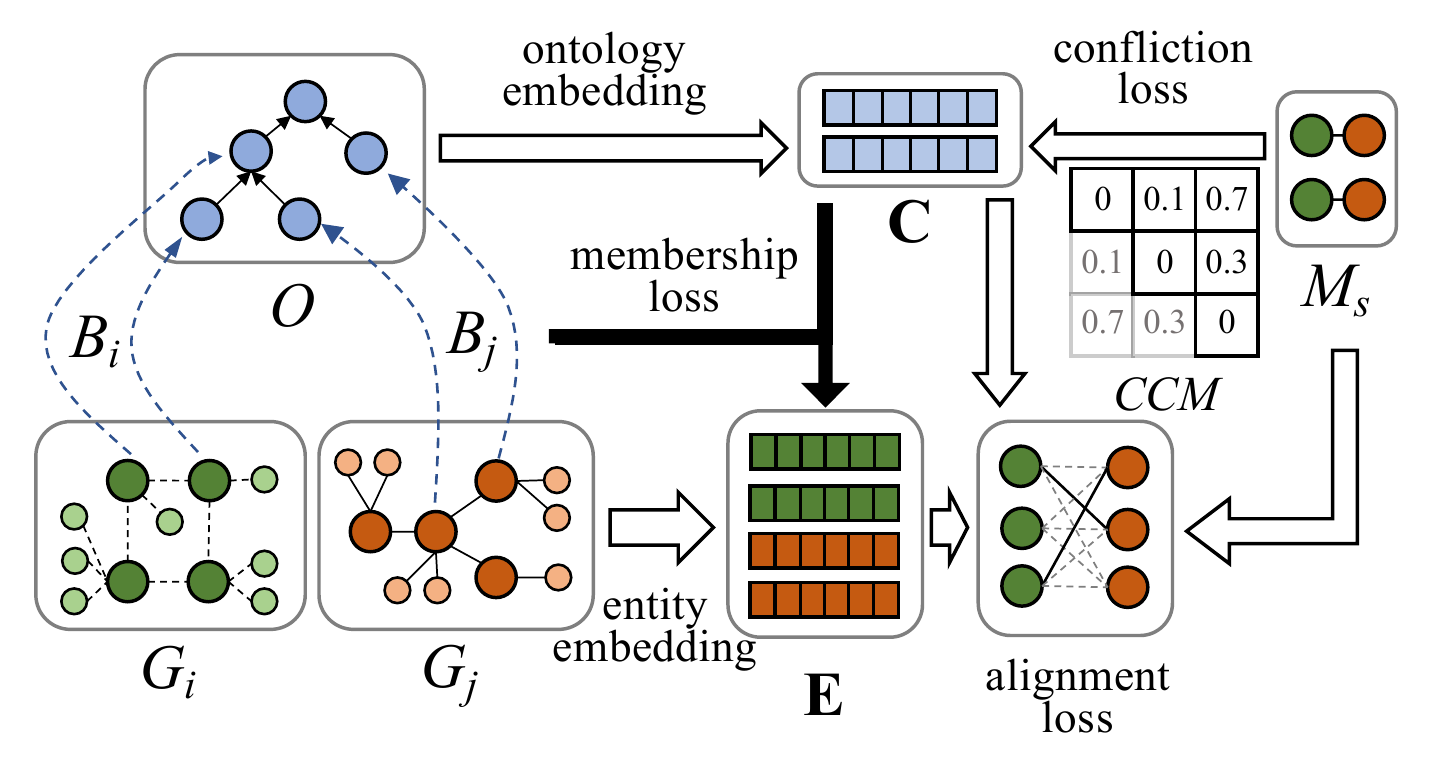}
\caption{The framework of \oea.
\label{figure:BriefFramework}}
\vspace{-5mm}
\end{figure}

\subsection{OntoEA Framework}

As shown in Fig.~\ref{figure:BriefFramework}, \oea includes five modules:
\textit{(i)} \textit{entity embedding} which embeds each KG into a separate embedding space;
\textit{(ii)} \textit{ontology embedding} which embeds the class hierarchical structure with a non-linear transformation;
\textit{(iii)} \textit{confliction loss} which incorporates all potential class conflicts in the embeddings with the CCM;
\textit{(iv)} \textit{membership loss} which incorporates the membership relationships and enables the joint learning of the entity embedding and the ontology embedding;
\textit{(v)} \textit{alignment loss} which bridges the embedding spaces of the to-be-aligned KGs via seed mappings.
An iterative co-training strategy is proposed to incorporate these five modules and the entity mappings are predicted with a pre-defined similarity measure between the joint embeddings.

\subsection{Entity Embedding}
\label{ent_embed}

To embed the KGs $G_i$ and $G_j$, we adopt the translation-based method TransE \cite{bordes2013translating} which interprets a triple $(h,r,t)$ as a translation by the relation $r$ from $h$ to $t$.
The margin-based loss of $(h,r,t)$ is defined as $f_{e}(h,t) = ||\bm{h}+\bm{r}-\bm{t}||_2$, where $||\cdot||_2$ denotes the L2 norm.
Besides, we extend the loss with a limit-based scoring loss \cite{zhou2017learning} to ensure the discrimination between the positive and negative triples and also lower scores for positive triples: 

\begin{equation}
\small
\begin{split}
    \mathcal{L}_E = \sum_{(h,t)\in T}\sum_{(h',t')\in T'}\{
    &[\gamma^1_e + f_{e}(h,t) - f_{e}(h',t')]_{+}\\
    &+\alpha_e[f_{e}(h,t) - \gamma^2_e]_{+}\},
\end{split}
\label{entity-loss}
\end{equation}
where $[\cdot]_{+}$ denotes the function $f(x)=\max(0,x)$, hyperparameters $\gamma^1_e$ and $\gamma^2_e$ control the margins,
$\alpha_e$ balances the margin-based loss and the limit-based loss,
and $T'$ is the set of the negative triples with each triple sampled using the $\epsilon$-truncated uniform negative sampling strategy~\cite{sun2018bootstrapping} to distinguish two similar triples.

It is worth noting that this study focuses on the EA task and the joint embedding challenge, and TransE is chosen due to its simplicity and efficiency.
Our \oea framework is open to other advanced KG embedding methods, such as GCN~\cite{wang2018cross} and RSN~\cite{guo2019learning}.

\subsection{Ontology Embedding}
\label{onto_embed}

The shared or merged ontology $O=(C,H)$ is composed of triples with the \textit{subClassOf} relation.
For simplicity, a triple $(c_h, r, c_t)$ is written as a class pair $(c_h, c_t)$ in $H$ where $r := subClassOf$.
Inspired by \citet{hao2019universal} on embedding the hierarchical graph structure, we calculate the scoring loss of $(c_h, c_t)$ with a non-linear transformation, $f_o(c_h,c_t) = ||\tanh(\bm{W}_o\bm{c}_h+\bm{b}_o)-\bm{c}_t||_2$, where $\bm{W}_o\in \mathcal{R}^{d_o\times d_o}$ and $\bm{b}_o\in\mathcal{R}^{d_o}$ are the learnable parameters, and $d_o$ denotes the ontology embedding dimension.
This tends to encode each class as a sphere and each subclass as a vector in the same semantic space after the non-linear transformation, and the relative positions are employed to model the relations between class and its subclass.

Similarly, we adopt the margin-based and limit-based loss for training:
\begin{equation}
\small
\begin{split}
    \mathcal{L}_O = \sum_{(c_h,c_t)\in H}\sum_{(c'_h,c'_t)\in H'}\{&[\gamma^1_o + f_o(c_h,c_t) - f_o(c'_h,c'_t)]_{+}\\
    &+\alpha_o[f_o(c_h,c_t) - \gamma^2_o]_{+}\},
\end{split}
\label{ontology-loss}
\end{equation}
where $H'$ denotes the set of negative class pairs with each pair sampled by replacing $c_h$ or $c_t$ following the uniform negative sampling strategy~\cite{bordes2013translating},
while $\gamma^1_o$, $\gamma^2_o$ and $\alpha_o$ are hyperparameters similar to $\gamma^1_e$, $\gamma^2_e$ and $\alpha_e$ in Eq.~\ref{entity-loss}.

Note that we do not directly utilize TransE in the ontology embedding, because the \textit{subClassOf} relation is transitive (e.g., we can infer (\textit{Royalty}, \textit{subClassOf}, \textit{Agent}) via (\textit{Royalty}, \textit{subClassOf}, \textit{Person}) and (\textit{Person}, \textit{subClassOf}, \textit{Agent}) as shown in Fig.~\ref{figure:intro}), which can lead to one-to-many and many-to-one mappings (or triples) in the ontology, and TransE cannot well address the relation with such transitive property \cite{lv2018differentiating}.

\subsection{Confliction Loss}
\label{conflict_embedding}

We use a class conflict matrix (CCM) to represent the inter-class conflicts that are either explicitly defined by class disjointness or implicitly discovered from the entities.
Within the CCM, the entry on the $i^{\text{th}}$ row and $j^{\text{th}}$ column, denoted as $m_{i,j}$, represents the conflict degree between class $c_i$ and class $c_j$.
For one ontology, the CCM is a squared and symmetric matrix, and we only maintain the upper triangular for higher efficiency.
Given two classes $c_i$ and $c_j$, $m_{i,j} \in \left[0,1\right]$ is calculated as follows.
First, we set $m_{i,j}=0$ if $c_i \equiv c_j$, which ensures each class does not conflict with itself.
Second, $c_i$ and $c_j$ are regarded as fully conflicted, i.e., $m_{i,j}=1$, if they are declared as disjointed by the ontology.\footnote{For ontologies by Web Ontology Language (OWL), the class disjointness is common and is usually defined as a constraint by the built-in relation \textit{owl:disjointWith}.}
Third, $c_i$ and $c_j$ are regarded as not conflicted, i.e., $m_{i,j}=0$, if they have at least one common member (entity) or some of their members are given as seed mapping (i.e., there exists $(e_i, e_j)$ in $M_s$ such that $e_i$ belongs to $c_i$ and $e_j$ belongs to $c_j$).
Note the above three conditions are matched sequentially, and the computation of $m_{i,j}$ is finished if any condition is met.
Finally, if none of the three conditions is met, we follow the principle that the farther two classes are separated in the tree-like class hierarchy structure, the lower semantic similarity and higher conflict degree the two classes have \cite{mumtaz2020frequency}.
To calculate the distance between $c_i$ and $c_j$, we use the set of classes passed by routing from $c_i$ and $c_j$ to the root class, denoted as $S(c_i)$ and $S(c_j)$, respectively, and then adopt the ratio of the intersection of $S(c_i)$ and $S(c_j)$.
Accordingly, we have $m_{i,j} = 1-\frac{|S(c_i) \cap S(c_j)|}{|S(c_i) \cup S(c_j)|}$ where $|\cdot|$ denotes the set cardinality.

For the implicitly discovered conflicts, we assume the small conflict degree between two classes if their embeddings are similar (i.e., high cosine similarity).
Thus we calculate another cosine similarity-based class conflict degree: $d_{cos}(c_i,c_j)=1-\cos(\bm{c}_i,\bm{c}_j)$, where $\bm{c}_i$ and $\bm{c}_j$ represent the embeddings of $c_i$ and $c_j$.
We propose to minimize the following negative log-likelihood loss to incorporate the class conflicts represented by CCM into the class embeddings,
\begin{equation}
    \small
    \mathcal{L}_C=-\sum_{c_i \in C}\sum_{c_j \in C}m_{i,j}\log d_{cos}(c_i,c_j).
    \label{ccm-loss}
\end{equation}

\subsection{Membership Loss}

We develop the membership embedding module to utilize the membership relationships, $B_i$ and $B_j$, to associate the KG embedding spaces with the ontology embedding space,
which is regarded as enhancing the KG embeddings with the ontology semantics.
Given one membership relationship $b=(e,c)$, we utilize a non-linear transformation to map the entity embeddings to the ontology embedding space, and we calculate the scoring loss as $f_{m}(e,c) = ||\tanh(\bm{W}_{m}\bm{e}+\bm{b}_{m})-\bm{c}||_2$ where $\bm{W}_m\in \mathcal{R}^{d_e\times d_o}$ and $\bm{b}_m\in\mathcal{R}^{d_o}$ are learnable parameters, $d_e$ and $d_o$ denote the dimension of KG embedding and ontology embedding, respectively.
Similarly, we minimize the following margin-based and limit-based loss to model all the membership relationships:
\begin{equation}
\small
\begin{split}
    \mathcal{L}_M = \sum_{(e,c)\in B}\sum_{(e',c')\in B'}\{&[\gamma^1_m + f_{m}(e,c) - f_{m}(e',c')]_{+}\\
    &+\alpha_m[f_{m}(e,c) - \gamma^2_m]_{+}\},
\end{split}
\label{member-loss}
\end{equation}
where $B'$ denotes the set of negative membership relationships with each relationship created by replacing the class $c$ following the uniform negative sampling strategy~\cite{bordes2013translating},
and $\gamma^1_m$, $\gamma^2_m$ and $\alpha_m$ are hyperparameters similar to $\gamma^1_e$, $\gamma^2_e$ and $\alpha_e$ in Eq.~\ref{entity-loss}.

\subsection{Alignment Loss}

For the alignment embedding module, \oea utilizes the seed mappings $M_s$ to bridge the embedding spaces of $G_i$ and $G_j$ such that the equivalent mappings between two cross-KG entities can be calculated via some distance metrics, such as cosine similarity.
Given a seed mapping $m=(e_i, e_j)$, its score is calculated as $f_{a}(e_i,e_j) = || \bm{W}_a\bm{e}_i-\bm{e}_j ||_2$ where $\bm{W}_a\in\mathcal{R}^{d_e\times d_e}$ is a learnable translation matrix, and the training loss is defined as,
\begin{equation}
\small
\mathcal{L}_{A} = \sum_{(e_i, e_j)\in M_s} f_{a}(e_i, e_j).
    \label{align-loss}
\end{equation}


\subsection{Iterative Co-Training and Prediction}
\label{sec:training}

To incorporate the aforementioned modules and obtain the embeddings of $G_i$, $G_j$ and $O$, we can directly minimize the following loss:
\begin{equation}
    \small
    \mathcal{L} = \mathcal{L}_{E} + \mathcal{L}_{O} + \lambda_{1}\mathcal{L}_{C} + \lambda_{2}\mathcal{L}_{M} + \lambda_{3}\mathcal{L}_{A},
    \label{all_loss}
\end{equation}
where $\lambda_1$, $\lambda_2$, and $\lambda_3$ are hyperparameters that balance the losses of confliction embedding, membership embedding and alignment embedding, respectively.
Instead of directly optimizing $\mathcal{L}$, we use an iterative co-training strategy in \oea to reduce model complexity and accelerate model convergence.
At each iteration, \oea first optimizes $\mathcal{L}_E$ and $\mathcal{L}_O$ independently, then sequentially optimizes $\mathcal{L}_C$ and $\mathcal{L}_M$, and finally optimizes $\mathcal{L}_A$. 
The iteration stops until some stopping criterion on the validation set is met.

With the embeddings of $G_i$, $G_j$ and $O$, we calculate the entity mappings with the cosine similarity.
Given two entities $e_i \in G_i$ and $e_j \in G_j$, the weighted similarity score is calculated as:
\begin{equation}
\small
    sim(e_i,e_j) = \beta \cos(\bm{e}_i,\bm{e}_j) + (1-\beta)\cos(\bm{c}_i,\bm{c}_j),
    \label{predict_eq}
\end{equation}
where hyperparameter $\beta \in \left[0,1\right]$ balances the similarities of entity embeddings and class embeddings.
It is possible that one entity has multiple classes declared; for example, \textit{Victoria} in Fig. \ref{figure:intro} can be an entity of \textit{Royalty} and another class like \textit{Female Leader}.
In this case we calculate the average of the embeddings of all the declared classes as the class embedding ($\bm{c}_i$ or $\bm{c}_j$).\footnote{Note that we do not consider the inferred classes, such as \textit{Person} for \textit{Victoria} in Fig. \ref{figure:intro}.}
In the prediction, for each entity in $G_i$ to be aligned,
we rank all candidate entities in $G_j$ by their weighted similarity scores.

\section{Benchmarks}

\subsection{Benchmark Details}

For comprehensive evaluation, we adopt six popular EA benchmarks including EN-FR-15K-V1/V2, EN-DE-15K-V1/V2 and D-W-15K-V1/V2 from ~\citet{sun2020benchmarking}, and an industrial benchmark MED-BBK-9K from~\citet{zhang2020industry}.
Note that for the cross-lingual benchmarks within DBpedia (i.e., EN-FR and EN-DE), the to-be-aligned KGs share one ontology, while for the other benchmarks, the to-be-aligned KGs have different ontologies that need to be aligned beforehand.

The benchmarks themselves do not include ontologies, and we thus extract and append an ontology for each KG, which includes the class structure (i.e., \textit{rdfs:subClassOf} relationships) and membership relationships.
Each benchmark therefore contains two KGs, their ontologies, and associated membership relationships.
The benchmark statistics are shown in Table~\ref{tab:dataset_stats} where ``-15K'' and ``9K'' in the benchmark names are omitted.
For the KGs of DBpedia, we use the DBpedia ontology and the membership relationships from the DBpedia SPARQL endpoint\footnote{http://dbpedia.org/sparql} by querying the classes of each entity with \textit{rdfs:type}.
Note we also utilize the defined class disjointness constraints by \textit{owl:disjointWith} for initializing CCM.
For the two Wikidata KGs in D-W-V1/V2, we extract their ontologies and membership relationships from the Wikidata SPARQL endpoint\footnote{https://query.wikidata.org/sparql} using queries with \textit{rdfs:subClassOf} and \textit{rdfs:type}.
For the KGs of the industrial benchmark, we use domain knowledge (with the assistance of medical experts) to construct a shared, small-scale, but high-quality ontology, and the corresponding membership relationships.


\begin{table}[t]
    \centering
    \footnotesize
    \renewcommand\arraystretch{1.15}
    \newcolumntype{P}[1]{>{\centering\arraybackslash}p{#1}}
    \caption{Statistics of the benchmarks.}
    \vspace{-0.2cm}
    \resizebox{1\columnwidth}{!}{
    \begin{threeparttable}
    \begin{tabular}{c|l|c|P{0.65cm}P{0.65cm}|P{0.85cm}P{0.8cm}}
        \Xhline{2\arrayrulewidth}
        \multicolumn{2}{c|}{ \multirow{2}*{Dataset}} & \multirow{2}{*}{KG} & \multicolumn{2}{c|}{Ontology\tnote{b}} & \multicolumn{2}{c}{Membership\tnote{c}} \\
        \multicolumn{2}{c|}{} & & \#Cls. & \#Trs. & \#Links & \#Roots \\
        \hline
         \parbox[t]{2mm}{\multirow{6}{*}{\rotatebox[origin=c]{90}{\textit{share-O\tnote{a}}}}} & EN-FR-V1 & EN/FR & 189 & 755 & 15,000 & 639 \\ 
         \cline{2-7}
         & EN-FR-V2 & EN/FR & 104 & 755 & 15,000 & 533 \\
         \cline{2-7}
         & EN-DE-V1 & EN/DE & 175 & 755 & 15,000 & 155 \\
         \cline{2-7}
         & EN-DE-V2 & EN/DE & 86 & 755 & 15,000 & 165 \\
         \cline{2-7}
         & \multirow{2}{*}{MED-BBK} & MED & 11 & 10 & 9,162 & 86 \\ 
         & & BBK & 11 & 10 & 9,162 & 3,362 \\ 
         \hline
         \parbox[t]{2mm}{\multirow{4}{*}{\rotatebox[origin=c]{90}{\textit{not-share-O}}}} & \multirow{2}{*}{D-W-V1} & DB & 172 & 755 & 15,000 & 306 \\ 
         & & WK & 140 & 695 & 15,000 & 342 \\ 
         \cline{2-7}
         & \multirow{2}{*}{D-W-V2} & DB & 71 & 755 & 15,000 & 463 \\ 
         & & WK & 68 & 695 & 15,000 & 418 \\ 
         \Xhline{2\arrayrulewidth}
    \end{tabular}
    \begin{tablenotes}
    \footnotesize
    \item[a]  ``\textit{share-O}'' means the to-be-aligned KGs share one ontology while ``\textit{not-share-O}'' means the to-be-aligned KGs have different ontologies.
    \item[b] ``\#Cls.'' denotes the number of classes and ``\#Trs.'' denotes the number of \textit{rdfs:subClassOf} relation triples.
    \item[c] ``\#Roots'' denotes the number of entities that have no \textit{rdf:type} property and are linked to the root class.
    \item[d] See Table \ref{tab:dataset_stats_more} in the Appendix for more benchmark statistics.
    \end{tablenotes}
    \end{threeparttable}
    }
    \label{tab:dataset_stats}
    \vspace{-5mm}
\end{table}

\subsection{Ontology Alignment}
\label{ontolgy_alignment_method}

For the benchmarks whose KGs do not share one ontology, we first align the ontologies.
We adopt two ontology alignment methods: \textit{manual annotation} and \textit{alignment system}.
In the first method, we employ five annotators to annotate class mappings for each ontology pair and the classes annotated by more than three annotators are adopted.
It is worth noting that the manual annotation of class mappings is worthwhile and often adopted in KG construction and curation because of the high quality and relatively small scale in comparison with the entity mappings.
In the second method, we apply a state-of-the-art ontology alignment system named PARIS\footnote{http://webdam.inria.fr/paris/} and some ad-hoc pre-processing and post-processing for automatic class mapping computation.
Please see Appendix~\ref{ontology_alignment_more} for more details on ontology alignment and merging.
For the D-W benchmarks, we considered both methods and compared their performance (see Table~\ref{tab:onto_alignment_analysis}).


\section{Experiments}

\begin{table*}[ht]
    \centering
    \footnotesize
    \renewcommand\arraystretch{1.15}
    \caption{Overall results of \oea and the baselines. 
    }
    \vspace{-0.2cm}
    \resizebox{1\textwidth}{!}{
    \begin{tabular}{l|l|ccc|ccc|ccc|ccc|ccc}
    \Xhline{2\arrayrulewidth}
         \multicolumn{2}{c}{ \multirow{2}*{Models}} & \multicolumn{3}{|c|}{EN-FR-15K-V1} & \multicolumn{3}{c|}{EN-FR-15K-V2} & \multicolumn{3}{c|}{D-W-15K-V1} & \multicolumn{3}{c|}{D-W-15K-V2} & \multicolumn{3}{c}{MED-BBK-9K} \\
         \multicolumn{2}{c|}{} & {\scriptsize H@1} & {\scriptsize H@5} & {\scriptsize MRR} & {\scriptsize H@1} & {\scriptsize H@5} & {\scriptsize MRR} & {\scriptsize H@1} & {\scriptsize H@5} & {\scriptsize MRR} & {\scriptsize H@1} & {\scriptsize H@5} & {\scriptsize MRR} & {\scriptsize H@1} & {\scriptsize H@5} & {\scriptsize MRR} \\
         \hline
         \parbox[t]{2mm}{\multirow{7}{*}{\rotatebox[origin=c]{90}{w/o SI}}} & MTransE & .247 & .467 & .351 & .240 & .436 & .336 & .259 & .461 & .354 & .271 & .490 & .376 & .004 & .014 & .012 \\
         & JAPE & .262 & .497 & .372 & .292 & .524 & .402 & .250 & .457 & .348 & .262 & .484 & .368 & .003 & .009 & .009 \\
         & SEA  & .280 & .530 & .397 & .360 & .651 & .494 & .360 & .572 & .458 & .567 & .770 & .660 & .199 & .375 & .287 \\
         & GCNAlign  & .338 & .589 & .451 & .414 & .698 & .542 & .364 & .580 & .461 & .506 & .743 & .612 & .065 & .153 & .117 \\
         & BootEA & \underline{.507} & \underline{.718} & \underline{.603} & \textbf{.660} & \underline{.850} & \underline{.745} & \underline{.572} & \underline{.744} & \underline{.649} & \textbf{.821} & \underline{.926} & \underline{.867} & \underline{.307} & \underline{.495} & \underline{.399} \\
         & RSN4EA & .393 & .595 & .487 & .579 & .759 & .662 & .441 & .615 & .521 & .723 & .854 & .782 & .195 & .311 & .253 \\
         & AliNet & .258 & .437 & .339 & .359 & .569 & .453 & .270 & .403 & .331 & .522 & .698 & .601 & .017 & .042 & .033 \\
         & \oea & \textbf{.566} & \textbf{.818} & \textbf{.678} & \underline{.654} & \textbf{.891} & \textbf{.757} & \textbf{.591} & \textbf{.808} & \textbf{.688} & \underline{.814} & \textbf{.950} & \textbf{.873} & \textbf{.343} & \textbf{.546} & \textbf{.440} \\
         \hline
         \multicolumn{2}{l|}{Improv. best \%} & 11.6 & 13.9 & 12.4 & -0.9 & 4.8 & 1.6 & 3.3 & 8.6 & 6.0 & -0.9 & 2.6 & 0.7 & 11.7 & 10.3 & 10.3 \\
         \Xhline{2\arrayrulewidth}
         \parbox[t]{2mm}{\multirow{4}{*}{\rotatebox[origin=c]{90}{w/ SI}}} & AttrE & .481 & .671 & .569 & .535 & .746 & .631 & .299 & .467 & .381 & .489 & .695 & .585 & .194 & .363 & .279 \\
         & MultiKE & .749 & .819 & .782 & \underline{.864} & .909 & \underline{.885} & .411 & .521 & .468 & .495 & .646 & .569 & .213 & .367 & .289 \\
         & RDGCN & \underline{.755} & \underline{.854} & \underline{.800} & .847 & \underline{.919} & .880 & \underline{.515} & \underline{.669} & \underline{.584} & \underline{.623} & \underline{.757} & \underline{.684} & \underline{.306} & \underline{.425} & \underline{.365} \\
         & \oea & \textbf{.797} & \textbf{.871} & \textbf{.832} & \textbf{.901} & \textbf{.981} & \textbf{.931} & \textbf{.553} & \textbf{.787} & \textbf{.656} & \textbf{.795} & \textbf{.943} & \textbf{.860} & \textbf{.517} & \textbf{.703} & \textbf{.604} \\
         \hline
         \multicolumn{2}{l|}{Improv. best \%} & 5.6 & 2.0 & 4.0 & 4.3 & 6.7 & 5.2 & 7.4 & 17.6 & 12.3 & 27.6 & 24.6 & 25.7 & 68.9 & 65.4 & 65.5  \\
    \Xhline{2\arrayrulewidth}
    \end{tabular}
    }
    \label{tab:main_result}
    \vspace{-2mm}
\end{table*}

\subsection{Experimental Setup}

We compare \oea against state-of-the-art EA models including four translation-based models, MTransE~\cite{chen2017multilingual};
JAPE~\cite{sun2017cross}; SEA~\cite{pei2019semi}; BootEA~\cite{sun2018bootstrapping}, two graph neural network based models, GCNAlign~\cite{wang2018cross}; AliNet~\cite{sun2020knowledge}, and one recurrent neural network based model, RSN4EA~\cite{guo2018recurrent}.
We also compare \oea against some state-of-the-art models that additionally utilize entity surface information (SI) (i.e., entity names) including AttrE~\cite{trisedya2019entity}, MultiKE~\cite{zhang2019MultiKE} and RDGCN~\cite{wu2019relation}.
Since using and not using SI are usually regarded as two different evaluation contexts~\cite{guo2019learning}, we separately evaluate \oea without (w/o) and with (w/) SI.
Note that \oea with SI is implemented by a simple but effective strategy which initializes the embeddings of the translation-based models in Section~\ref{ent_embed} and Section~\ref{onto_embed} (i.e., $\bm{h}$, $\bm{r}$, $\bm{t}$ and $\bm{c}$) with the pre-trained word embeddings of their names.
As in~\citet{sun2020benchmarking}, we use the unified multi-lingual word embeddings fastText~\cite{bojanowski2017enriching} for the cross-lingual benchmarks.\footnote{The word embeddings are publicly available at https://fasttext.cc/docs/en/crawl-vectors.html.}
All the results of AliNet, and the results of MTransE, JAEP, SEA and GCNAlign on MED-BBK-9K are reproduced locally; while the other baseline results are taken from \citet{sun2020benchmarking} and \citet{zhang2020industry}.
We follow the same train ($20\%$), validation ($10\%$) and test ($70\%$) splits as \citet{sun2020benchmarking} and \citet{zhang2020industry}.


We implement \oea upon the open source library OpenEA\footnote{https://github.com/nju-websoft/OpenEA}.
We use the AdaGrad optimizer with a learning rate of $0.01$. 
The batch sizes for entity embedding and ontology embedding are set to 4500 and 64, respectively, and their dimensions are both set to 300.
The weight hyperparameters are set as $\lambda_1=1$, $\lambda_2=1$, $\lambda_3=5$ and $\beta=0.5$, and other hyperparameters are set as $\gamma^1_{x}=0.01$, $\gamma^2_{x}=2.0$ and $\alpha_{x}=0.2$ for all the losses where $x\in\{e,o,m\}$.
These hyperparamters are tuned w.r.t. the MRR on the validation set.
Please see Appendix~\ref{more_implementation_detail} for more implementation details.

With the embeddings we use the nearest neighbor search with cross-domain similarity local scaling \cite{lample2018word} to calculate the entity matching of each to-be-aligned entity.
In the evaluation, we rank matching candidates of each to-be-aligned entity and calculate the metrics of Hits@1 (H@1), Hits@5 (H@5) and mean reciprocal rank (MRR).
In the following tables, the best (second best resp.) result is \textbf{bolded} (\underline{underlined} resp.).

\begin{figure*}[ht]
    \centering
    \includegraphics[width=\linewidth]{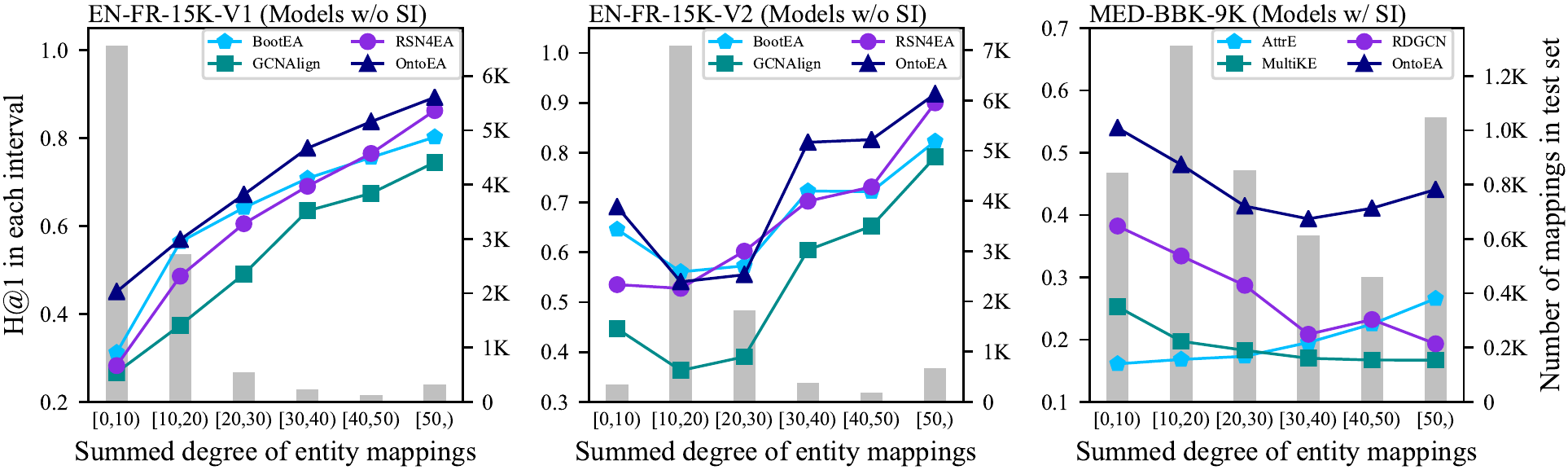}
    \caption{Results of \oea and some baselines on test entity mappings with different summed degrees.
    }
    \label{fig:long_tail_study}
    \vspace{-2mm}
\end{figure*}

\subsection{Overall Results}
Table~\ref{tab:main_result} reports the overall results of \oea and the baselines.\footnote{Results on the two EN-DE benchmarks, which lead to similar findings as the EN-FR benchmarks, are shown in Table~\ref{tab:main_result_more} in the Appendix.}
Overall, in both contexts of using and not using SI, \oea performs the best across all the benchmarks on all the metrics, except for H@1 on EN-DE-15K-V2 and D-W-15K-V2.

\noindent
\textbf{Without SI.} 
Regarding the benchmarks that share one ontology, \oea achieves at least $10\%$ higher performance in all the metrics over the best baseline BootEA on EN-FR-15K-V1 which has sparse KG structures.
On EN-FR-15K-V2 with dense KG structures, \oea achieves higher H@5 and MRR but competitive H@1 in comparison with the best baseline.
For the D-W benchmarks whose KGs have different ontologies, we have a similar finding: the outperformance of \oea is more significant on the benchmark with sparse KG structures (i.e., D-W-15K-V1) than on the benchmark with dense KG structures (i.e., D-W-15K-V2).
On the one hand, dense KG structures indicate that more information can be utilized with better performance, and relatively the additional positive impact by the ontology becomes more limited.
On the other hand, the ontologies of both EN-FR-15K-V2 and D-W-15K-V2 coincidentally have fewer classes (see Table \ref{tab:dataset_stats}), which would also lead to a lower impact of the ontology guidance.
Regarding the industrial MED-BBK-9K, which is quite new and has shown more challenging w.r.t. these baselines \cite{zhang2020industry}, \oea outperforms the best baseline by more than 10\% in all the metrics.
Although its ontology has a limited scale, the ontology is of high quality as it is specifically created with the domain knowledge.

\noindent
\textbf{With SI.} 
\oea shows very promising overall results as \oea without SI. It outperforms all the baselines on all the benchmarks, with the improvements over the best baseline ranging from $2\%$ to $68.9\%$.
Regarding the industrial MED-BBK-9K, the performance improvement of \oea is especially significant with more than 60\% in all the metrics.
As aforementioned, this benchmark is challenging to these baselines and it has high-quality ontologies.
It is worth mentioning that on the D-W benchmarks, the involvement of SI deteriorates the performance of \oea and the best baseline,
while the positive impact of the ontologies (i.e., the performance gain of \oea over the best baseline) becomes more significant.

\subsection{Ablation Studies}


\noindent
\textbf{Model Component Analysis.} 
To investigate different components in \oea, we compare \oea variants without the CCM loss (w/o $\mathcal{L}_C$), without the membership relationship loss (w/o $\mathcal{L}_M$), and without the ontology (w/o Onto.).
The experimental results of \oea w/o SI on EN-FR-15K-V1/V2 are reported in Table~\ref{tab:ablation_study}, which shows that $\mathcal{L}_M$ contributes more to \oea than $\mathcal{L}_C$ as removing it leads to a larger performance drop.
The results \oea w/ SI show similar findings and are thus omitted.
As expected, the removal of the ontology deteriorates performance the most.
All these results verify the significant role of the ontologies and their associated class conflicts and membership relationships.


\begin{table}[ht]
    \centering
    \small
    \renewcommand\arraystretch{1.15}
    \caption{Results of \oea w/o SI and its variants with the removal of different components.}
    \vspace{-0.2cm}
    \resizebox{1.0\linewidth}{!}{
    \begin{tabular}{l|ccc|ccc}
    \Xhline{2\arrayrulewidth}
    \multirow{2}{*}{Models} & \multicolumn{3}{c|}{EN-FR-15K-V1} & \multicolumn{3}{c}{EN-FR-15K-V2} \\
    & {\scriptsize H@1} & {\scriptsize H@5} & {\scriptsize MRR} & {\scriptsize H@1} & {\scriptsize H@5} & {\scriptsize MRR} \\
    \hline
    \oea & \textbf{.566} & \textbf{.818} & \textbf{.678} & \textbf{.654} & \textbf{.891} & \textbf{.757} \\
    w/o $\mathcal{L}_{C}$ & \underline{.520} & \underline{.781} & \underline{.636} & \underline{.589} & \underline{.845} & \underline{.701}  \\
    w/o $\mathcal{L}_{M}$ & .481 & .750 & .601 & .549 & .810 & .665 \\
    w/o Onto. & .430 & .698 & .551 & .545 & .814 & .664 \\ 
    \Xhline{2\arrayrulewidth}
    \end{tabular}
    }
    \label{tab:ablation_study}
    \vspace{-1mm}
\end{table}

\noindent
\textbf{Results on Different Test Entity Mappings.}
We split the test mappings according to the summed degree, i.e., $\text{deg}(e_i, e_j) := \text{deg}(e_i) + \text{deg}(e_j)$ where $(e_i,e_j)$ are the entity mappings.
We select three different experimental settings (benchmarks and usage of SI) and Fig.~\ref{fig:long_tail_study} shows the results of \oea and some competitive baselines on different degree intervals.
On EN-FR-15K-V1, \oea outperforms the baselines on all the intervals and the performance gap reaches the highest on [0,10).
On EN-FR-15K-V2 with dense KG structures, \oea performs close to or slightly worse than BootEA on intervals [10,20) and [20,30) but outperforms all the baselines on the other intervals.
On MED-BBK-9K, \oea performs much better than the baselines on all the intervals.
All these observations are consistent with our findings from the overall results in Table~\ref{tab:main_result} and also verify the effectiveness of \oea on different test entity mappings.


\begin{figure}[ht]
    \centering
    \includegraphics[width=\linewidth]{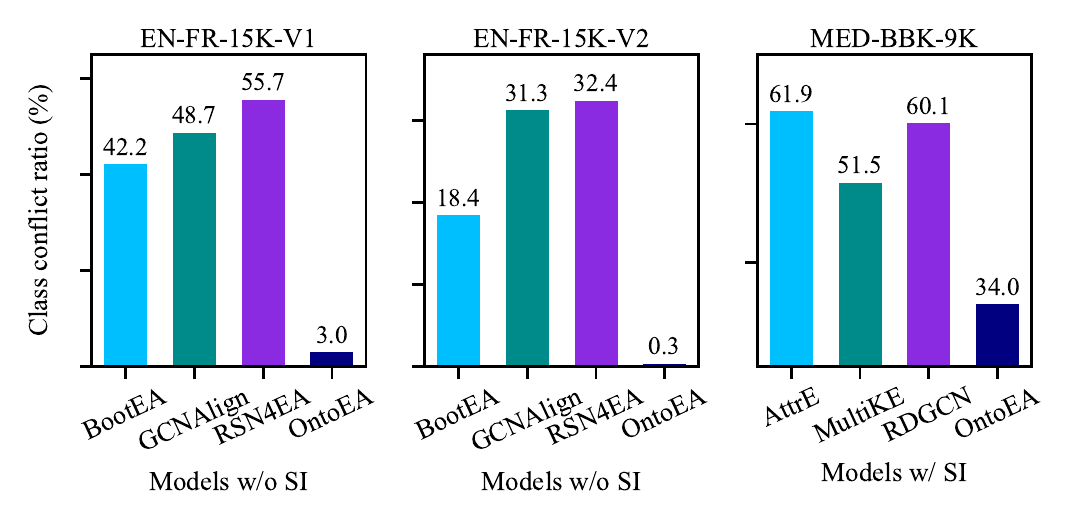}
    \caption{The class conflict ratios.}
    \label{fig:conflict_ratio_study}
    \vspace{-4mm}
\end{figure}

\noindent
\textbf{Analysis of Class Conflicts.}
This part analyses the predicted mappings that have class conflicts, with the results shown in Fig.~\ref{fig:conflict_ratio_study}. Note that the class conflict ratio of a method is the rate of its false positive mappings with class conflicts among all its false positive mappings.
On EN-FR-15K-V1 and EN-FR-15K-V2, \oea significantly decreases the class conflict ratio to $3\%$ and $0.3\%$, respectively, much lower than those of the baselines.
On MED-BBK-9K, \oea reduces the class conflict ratio to $34.0\%$ while MultiKE and RDGCN have $51.5\%$ and $60.1\%$, respectively.
The above observations illustrate that \oea has effectively implemented our motivation of using the ontologies and class conflicts to avoid false positive mappings.

\noindent
\textbf{Analysis of Ontology Alignment Methods}. 
As the case that the to-be-aligned KGs have different ontologies is common, we analyse the impact of different ontology alignment methods proposed in Sec.~\ref{ontolgy_alignment_method}, with the results in Table~\ref{tab:onto_alignment_analysis}.
We find that \oea with manual annotation achieves better results than \oea using the PARIS based alignment system method;
the latter, however, still has competitive or better results in comparison with the best baseline (i.e., BootEA or RDGCN).
This also provides motivations for the future research and industrial deployment work on iterative ontology alignment to keep a balance between the annotation cost and the alignment quality.

\begin{table}[ht]
    \centering
    \small
    \renewcommand\arraystretch{1.15}
    \caption{Results of \oea using different ontology alignment methods where ``-M'' denotes the manual annotation while ``-A'' denotes the alignment system.}
    \vspace{-0.2cm}
    \resizebox{1.0\linewidth}{!}{
    \begin{tabular}{l|l|ccc|ccc}
    \Xhline{2\arrayrulewidth}
    & \multirow{2}{*}{Models} & \multicolumn{3}{c|}{D-W-15K-V1} & \multicolumn{3}{c}{D-W-15K-V2} \\
    & & {\scriptsize H@1} & {\scriptsize H@5} & {\scriptsize MRR} & {\scriptsize H@1} & {\scriptsize H@5} & {\scriptsize MRR} \\
    \hline
    \parbox[t]{2mm}{\multirow{3}{*}{\rotatebox[origin=c]{90}{w/o SI}}} & \oea-M & \textbf{.591} & \textbf{.808} & \textbf{.688} & \underline{.814} & \textbf{.950} & \textbf{.873} \\
    & \oea-A & \underline{.572} & \underline{.795} & \underline{.673} & .762 & .907 & .827 \\
    & \textit{BootEA} & .572 & .744 & .649 & \textbf{.821} & \underline{.926} & \underline{.867} \\
    \hline
    \parbox[t]{2mm}{\multirow{3}{*}{\rotatebox[origin=c]{90}{w/ SI}}} & \oea-M & \textbf{.553} & \textbf{.787} & \textbf{.656} & \textbf{.795} & \textbf{.943} & \textbf{.860} \\
    & \oea-A & .510 & \underline{.758} & \underline{.621} & \underline{.737} & \underline{.893} & \underline{.808} \\
    & \textit{RDGCN} & \underline{.515} & .669 & .584 & .623 & .757 & .684 \\
    \Xhline{2\arrayrulewidth}
    \end{tabular}
    }
    \label{tab:onto_alignment_analysis}
    \vspace{-2mm}
\end{table}

\section{Related Work}


\noindent
\textbf{Embedding-based KG Alignment.} 
These methods can be categorised into translation-based,  GNN-based and  RNN-based.
The translation-based methods mainly rely upon some translation-based KG embedding models; for example, MTransE adopts TransE and various mapping losses to align cross-lingual KGs~\cite{chen2017multilingual}.
To utilize more useful information in the KG, JAPE embeds the to-be-aligned KGs into one unified vector space via leveraging attribute correlations~\cite{sun2017cross}; AttrE exploits attributes by generating attribute character embeddings~\cite{trisedya2019entity}; and MultiKE makes full use of entity names, entity attributes and graph structures via multi-view learning~\cite{zhang2019MultiKE}.
The GNN-based methods are typically based on GNN variants for KG embedding, such as Graph Convolutional Networks (GCNs) \citet{wang2018cross}, dual-primal GCNs \citet{sun2020knowledge}, and GCNs with attentive aggregation \citet{sun2020knowledge}.
As an RNN-based method, RSN4EA generates biased random walks on KGs and learns the embeddings by a recurrent skipping network~\cite{guo2019learning}.
Some work has explored a semi-supervised learning setting, and for instance, BootEA~\cite{sun2018bootstrapping} adopts the bootstrapping strategy to iteratively append new likely mappings during the learning process.

\noindent
\textbf{KG Embedding with Ontology.}
Only a few attempts have been made towards the KG embedding with ontology.
The most relevant work is JOIE \cite{hao2019universal} which embeds the KG and the ontology in two embedding spaces and enables them to enhance each other by the cross-view links i.e., membership relationships.
Some ontology embedding methods such as OWL2Vec*~\cite{chen2020owl2vec} can be extended to jointly embed KG plus ontology since the KG can be regarded as an assertion part (ABox) of the ontology.
However, as far as we know, there are currently no embedding methods that support two KGs along with their ontologies, let alone utilize the ontologies to augment embedding-based KG alignment.

\section{Conclusions}
This paper presented a novel method \oea to augment embedding-based entity alignment with the ontology.
\oea enriches the semantics of KG embeddings by jointly learning the KGs, the ontology, the membership relationships and the seed mappings, and utilizes the potential class conflicts to avoid false mappings.
The evaluation on multiple benchmarks has showed that \oea could achieve state-of-the-art performance and reduce the false positive mappings with class conflicts.
For future work, we plan to utilize other ontology semantics besides the class hierarchy and class disjointness constraints~\cite{chen2021augmenting}.
We also plan to extend the entity embedding module with other advanced KG embedding methods.


\bibliographystyle{acl_natbib}
\newpage
\clearpage
\balance
\bibliography{ref-new}

\newpage
\clearpage
\appendix

\section*{Appendix}
\label{sec:appendix}

\section{More Benchmark Details}

Table~\ref{tab:dataset_stats_more} shows the full statistics of all the benchmarks used in the experiments, including four cross-lingual benchmarks and two cross-KG benchmarks from OpenEA~\cite{sun2020benchmarking},
and one recent industrial cross-KG benchmark from~\citet{zhang2020industry}. We report the benchmark statistics from three perspectives: the ontologies, the KGs and the membership relationships.
The benchmarks are divided as previous work that 20\%, 10\% and 70\% of all the entity mappings are used as training, validation and test sets, respectively~\cite{sun2020benchmarking,zhang2020industry}.

\section{More Implementation Details}
\label{more_implementation_detail}
Our experiments are conducted on a workstation with an Intel Xeon E5 2.40 GHz CPU, 128 GB memory, an NVIDIA Tesla M40 GPU, and CentOS 7.2. 
Search spaces of the hyperparameters used in \oea are listed in Table~\ref{tab:hyperparameters}.
We apply a grid search strategy for the hyperparameter configuration that performs the best on the validation set with respect to mean reciprocal rank (MRR).

\begin{table}[ht]
    \small
    \centering
    \caption{Search spaces of the hyperparameters.}
    \begin{threeparttable}
    \begin{tabular}{p{2.8cm}r}
        \Xhline{2\arrayrulewidth}
        Hyperparameters & Search spaces  \\
        \hline
        learning rate & \{1e-3, 5e-3, 1e-2, 5e-2\} \\
        batch size (\textit{entity}) & \{4000, 4500, 5000\} \\
        batch size (\textit{ontology}) & \{32, 64, 128\} \\
        $\gamma^1_{x}$ in Eqs. (\ref{entity-loss}) (\ref{ontology-loss}) (\ref{member-loss}) \tnote{*} & \{0.01, 0.02, 0.03\} \\
        $\gamma^2_{x}$ in Eqs. (\ref{entity-loss}) (\ref{ontology-loss}) (\ref{member-loss}) \tnote{*} & \{1.0, 2.0, 3.0\} \\
        $\alpha_x$ in Eqs. (\ref{entity-loss}) (\ref{ontology-loss}) (\ref{member-loss}) \tnote{*} & \{0.1, 0.2, 0.3\} \\
        $\lambda_1$ in Eq. (\ref{all_loss}) & \{0, 1, 2, 3, 4, 5\} \\
        $\lambda_2$ in Eq. (\ref{all_loss}) & \{0, 1, 2, 3, 4, 5\} \\
        $\lambda_3$ in Eq. (\ref{all_loss}) & \{0, 1, 2, 3, 4, 5\} \\
        $\beta$ in Eq. (\ref{predict_eq}) & \{0.3, 0.4, 0.5, 0.6, 0.7\} \\
        \Xhline{2\arrayrulewidth}
    \end{tabular}
    \begin{tablenotes}
    \item[*]where $x\in\{e,o,m\}$.
    \end{tablenotes}
    \end{threeparttable}
    \label{tab:hyperparameters}
\end{table}

\begin{table*}[ht]
    \centering
    \footnotesize
    \renewcommand\arraystretch{1.15}
    \caption{Full statistics of all the benchmarks.}
    \vspace{-0.2cm}
    \begin{threeparttable}
    \begin{tabular}{c|c|c|cc|cccc|cc}
        \Xhline{2\arrayrulewidth}
        \multicolumn{2}{c|}{ \multirow{2}*{Dataset} } & \multirow{2}{*}{KG} & \multicolumn{2}{c|}{Ontology \tnote{b}} & \multicolumn{4}{c|}{Knowledge Graph} & \multicolumn{2}{c}{Membership \tnote{c}} \\
        \multicolumn{2}{c|}{} & & \#Cls. & \#Sub tr. & \#Rel. & \#Attr. & \#Rel tr. & \#Attr tr. & \#Links & \#Roots \\
        \hline
         \parbox[t]{2mm}{\multirow{10}{*}{\rotatebox[origin=c]{90}{\textit{share-O \tnote{a}}}}} & \multirow{2}{*}{EN-FR-15K-V1} & EN & \multirow{2}{*}{189} & \multirow{2}{*}{755} & 267 & 308 & 47,334 & 73,121 & \multirow{2}{*}{15,000} & \multirow{2}{*}{639} \\ 
         & & FR & & & 210 & 404 & 40,864 & 67,167 & & \\ 
         \cline{2-11}
         & \multirow{2}{*}{EN-FR-15K-V2} & EN & \multirow{2}{*}{104} & \multirow{2}{*}{755} & 193 & 189 & 96,318 & 66,899 & \multirow{2}{*}{15,000} & \multirow{2}{*}{533} \\ 
         & & FR & & & 166 & 221 & 80,112 & 68,779 & & \\ 
         \cline{2-11}
         & \multirow{2}{*}{EN-DE-15K-V1} & EN & \multirow{2}{*}{175} & \multirow{2}{*}{755} & 215 & 286 & 47,676 & 83,775 & \multirow{2}{*}{15,000} & \multirow{2}{*}{155} \\ 
         & & DE & & & 131 & 194 & 50,419 & 156,150 & & \\ 
         \cline{2-11}
         & \multirow{2}{*}{EN-DE-15K-V2} & EN & \multirow{2}{*}{86} & \multirow{2}{*}{755} & 169 & 171 & 84,867 & 81,998 & \multirow{2}{*}{15,000} & \multirow{2}{*}{165} \\ 
         & & DE & & & 96 & 116 & 92,632 & 186,335 & & \\ 
         \cline{2-11}
         & \multirow{2}{*}{MED-BBK-9K} & MED & 11 & 10 & 32 & 19 & 158,357 & 11,467 & 9,162 & 86 \\ 
         & & BBK & 11 & 10 & 20 & 21 & 50,307 & 44,987 & 9,162 & 3,362 \\ 
         \hline
         \parbox[t]{2mm}{\multirow{4}{*}{\rotatebox[origin=c]{90}{\textit{not-share-O}}}} & \multirow{2}{*}{D-W-15K-V1} & DB & 172 & 755 & 248 & 342 & 38,265 & 68,258 & 15,000 & 306 \\ 
         & & WK & 140 & 695 & 269 & 649 & 42,746 & 138,246 & 15,000 & 342 \\ 
         \cline{2-11}
         & \multirow{2}{*}{D-W-15K-V2} & DB & 71 & 755 & 167 & 175 & 73,983 & 66,813 & 15,000 & 463 \\ 
         & & WK & 68 & 695 & 121 & 457 & 83,365 & 175,686 & 15,000 & 418 \\ 
         \Xhline{2\arrayrulewidth}
    \end{tabular}
    \begin{tablenotes}
    \footnotesize
    \item[a]  ``\textit{share-O}'' means the to-be-aligned KGs share one ontology while ``\textit{not-share-O}'' means the to-be-aligned KGs have different ontologies.
    \item[b] ``\#Cls.'' denotes the number of classes and ``\#Trs.'' denotes the number of \textit{rdfs:subClassOf} relation triples.
    \item[c] ``\#Roots'' denotes the number of entities that have no \textit{rdf:type} property and are linked to the root class.
    \end{tablenotes}
    \end{threeparttable}
    \label{tab:dataset_stats_more}
\end{table*}

\section{Ontology Alignment Methods}
\label{ontology_alignment_more}
In this section, we present the details of the two ontology alignment methods: \textit{manual annotation} and \textit{alignment system}.

\subsection{Manual Annotation}

We refer to crowdsourcing for manual annotation of the class mappings by the following steps which are presented with the example of the DBpedia ontology and the Wikidata ontology in the D-W benchmarks.
First, we recruit five volunteers with at least undergraduate education as the annotators and provide them the two to-be-aligned ontologies.
The DBpedia ontology ($O_D$) contains a total of $755$ classes while the Wikidata ontology ($O_W$) contains $695$ classes.
Second, for each class in $O_D$, denoted as $c_D$, each annotator is required to consider as more classes in $O_W$ as possible and find the equivalent class mapping $(c_D,c_W)$, where $c_W$ denotes a class in $O_W$.
Note that the annotators are allowed to utilize her/his knowledge and accelerate the annotation by searching for class $c_W \in O_W$, using some friendly ontology management and accessing software such as Prot\'eg\'e. 
After all five annotators finish the work, only the class mappings that are labelled by more than three annotators are accepted as the final class mappings, denoted as $M_C$.

With the class mappings, we need to further merge the two ontologies as one shared ontology which can then be processed by \oea. 
To this end, our solution is to build new membership relationships between the entities of the Wikidata KG and the classes of the DBpedia ontology.
For each original membership relationship within the Wikidata KG and ontology, we replace its class $c_W$ with $c_D$ if $(c_D,c_W) \in M_C$, or with the root class \textit{owl:Thing} otherwise.

\subsection{Alignment System}

\begin{algorithm}
\caption{Alignment System Method}
\begin{algorithmic}[1]
\small
\REQUIRE Ontologies $O_D$ and $O_W$, threshold $\lambda$
\STATE Build directed graphs $D_D$ and $D_W$ from $O_D$ and $O_W$, respectively, with the relation of \textit{rdfs:subClassOf}
\STATE Obtain equivalent class pairs $P_C$ from pre-defined  \textit{owl:equivalentClass} in $O_D$ and $O_W$
\STATE Generate $D'$ via connecting $D_D$ and $D_W$ with $P_C$
\STATE Define $\phi(c_1,c_2)$ as the length of directed walk from $c_1$ to $c_2$ where $c_1,c_2 \in D'$
\STATE Initialize empty mapping sets $M_C$ and $M'_C$
\FOR{each class $c_W$ in $O_W$}
    \STATE Route $D'$ starting from $c_W$ and output routed class set $S=\{c\}$ where $c \in D'$
    \IF{$S = \emptyset$}
    \STATE Append $(c_W,$\textit{owl:Thing}$)$ to $M_C$
    \ELSE
    \STATE Find a class $c$ in $S$ such that for any $c'$ in $S$ $\phi(c,$\textit{owl:Thing}$) \geq \phi(c',$\textit{owl:Thing}$)$
    \STATE Append $(c_W,c)$ to $M_C$
    \ENDIF
\ENDFOR
\STATE Perform PARIS on $O_D$ and $O_W$
\STATE Append $M'_C$ from PARIS with threshold $\lambda$
\FOR{each class mapping $(c_w,c_d)$ in $M_C$}
    \IF{$(c_w,c'_d) \in M'_C$ and $c'_d \neq c_d$}
    \STATE Remove $(c_w,c_d)$ from $M_C$ 
    \STATE Append $(c_w,c'_d)$ to $M_C$
    \ENDIF
\ENDFOR
\RETURN $M_C$
\end{algorithmic}
\label{algo}
\end{algorithm}

The automatic method based on the ontology alignment system PARIS and some ad-hoc pre-processing and post-processing is shown in Algorithm~\ref{algo}.
The pre-processing from Line 1 to Line 13 obtains class mappings via finding the most fine-grained mappings (i.e., the farthest class node in the directed graph).
Since PARIS generally outputs reliable class mappings, the post-processing from Line 16 to Line 21 updates class mappings with those output from PARIS.

In the implementation we set PARIS hyperparameter $\lambda$ to $0.4$.
The \textit{owl:equivalentClass} relationship across the DBpedia ontology and the Wikidata ontology is acquired via DBpedia and Wikidata SPARQL endpoints. 
After getting $M_C$, we merge the two ontologies as introduced in the method of manual annotation.

\section{More Experimental Results}

\begin{table}[ht]
    \centering
    \small
    \renewcommand\arraystretch{1.15}
    \caption{Overall results of \oea and the baselines on the EN-DE benchmarks.}
    \vspace{-0.2cm}
    \resizebox{1\linewidth}{!}{
    \begin{tabular}{l|l|ccc|ccc}
    \Xhline{2\arrayrulewidth}
         \multicolumn{2}{c}{\multirow{2}*{Models}} & \multicolumn{3}{|c|}{EN-DE-15K-V1} & \multicolumn{3}{c}{EN-DE-15K-V2} \\
         \multicolumn{2}{c|}{} & {\scriptsize H@1} & {\scriptsize H@5} & {\scriptsize MRR} & {\scriptsize H@1} & {\scriptsize H@5} & {\scriptsize MRR} \\
         \hline
         \parbox[t]{2mm}{\multirow{7}{*}{\rotatebox[origin=c]{90}{w/o SI}}} & MTransE & .307 & .518 & .407 & .193 & .352 & .274 \\
         & JAPE & .288 & .512 & .394 & .167 & .329 & .250  \\
         & SEA & .530 & .718 & .617 & .606 & .779 & .687 \\
         & GCNAlign & .481 & .679 & .571 & .534 & .717 & .618 \\
         & BootEA & \underline{.675} & \underline{.820} & \underline{.740} & \underline{.833} & \underline{.912} & \underline{.869} \\
         & RSN4EA & .587 & .752 & .662 & .791 & .890 & .837 \\
         & AliNet & .243 & .353 & .295 & .169 & .247 & .277 \\
         & OntoEA & \textbf{.742} & \textbf{.897} & \textbf{.812} & \textbf{.864} & \textbf{.955} & \textbf{.905} \\
         \hline
         \multicolumn{2}{l|}{Improv. best \%} & 9.9 & 9.4 & 9.7 & 3.7 & 4.7 & 4.1 \\
         \Xhline{2\arrayrulewidth}
         \parbox[t]{2mm}{\multirow{4}{*}{\rotatebox[origin=c]{90}{w/ SI}}} & AttrE & .517 & .687 & .597 & .650 & .816 & .726 \\
         & MultiKE & .756 & .809 & .782 & .755 & .813 & .784 \\
         & RDGCN & \underline{.830} & \underline{.895} & \underline{.859} & \underline{.833} & \underline{.891} & \underline{.860} \\
         & OntoEA & \textbf{.850} & \textbf{.906} & \textbf{.877} & \textbf{.914} & \textbf{.977} & \textbf{.942} \\
         \hline
         \multicolumn{2}{l|}{Improv. best \%} & 2.4 & 1.2 & 2.1 & 9.7 & 9.7 & 9.5 \\
    \Xhline{2\arrayrulewidth}
    \end{tabular}
    }
    \label{tab:main_result_more}
\end{table}

Table~\ref{tab:main_result_more} reports the overall results on the cross-lingual benchmarks  EN-DE-15K-V1 and EN-DE-15K-V2. 
We have similar findings in these two EN-DE benchmarks as in the EN-FR benchmarks that \oea consistently outperform baselines, w/o SI and w/ SI.
On EN-DE-15K-V1, \oea generally outperforms competitive models by up to 9.9\% in H@1 and the performance gap is larger in the models w/o SI than in the models w/ SI.
On the dense benchmark (V2), \oea shows limited performance gain (about 4\%) compared with the models w/o SI but much larger performance gain (more than 9\%) compared with the models w/ SI.

\end{document}